\documentclass{article}
\usepackage{graphicx}
\usepackage{multirow}
\usepackage[nice]{nicefrac}
\usepackage{amsmath}

\newcommand{\ccirc}{\kern0.5ex\vcenter{\hbox{$\scriptstyle\circ$}}\kern0.5ex}
\def\LogScale{L}
\def\UniformScale{U}
\usepackage[usenames,dvipsnames,svgnames,table]{xcolor}
\usepackage[colorlinks=true,allcolors=RoyalPurple]{hyperref}
\usepackage{natbib}
\title{Low-Precision Batch-Normalized Activations}
\author{
  Benjamin Graham
  \\
  Facebook AI Research
  \\
  \texttt{benjamingraham@fb.com}
}

\begin{document}
\maketitle
\begin{abstract}
Artificial neural networks can be trained with relatively low-precision floating-point and fixed-point arithmetic, using between one and 16 bits. Previous works have focused on relatively wide-but-shallow, feed-forward networks. We introduce a quantization scheme that is compatible with training very deep neural networks. Quantizing the network activations in the middle of each batch-normalization module can greatly reduce the amount of memory and computational power needed, with little loss in accuracy.
\end{abstract}

\section{Introduction}

To improve accuracy, deeper and deeper neural networks are being trained,
leading to seemingly ever-improving performance in computer vision,
natural language processing, and speech recognition.
Three techniques have been key to usefully increasing the depth of networks:
rectified-linear units (ReLU) \citep{conf/icml/NairH10},
batch normalization \citep{BatchNorm}, and residual connections \citep{ResNet,DBLP:journals/corr/WuSCLNMKCGMKSJL16,journals/corr/SzegedyIV16,DenseNet,Zagoruyko2016WRN,larsson2016fractalnet,oai:arXiv.org:1602.07360}.

These techniques have fulfilled the promise of deep learning,
allowing networks with a large number of relatively narrow layers
to achieve high accuracy. The experiments that established the usefulness of these techniques were carried out using
32-bit precision floating-point arithmetic.

Network architectures are often compared in terms of the number of
parameters in the network, and the number of FLOPs per test evaluation.
Memory requirements, either at training time or test time are rarely
explicitly taken into account, even though they often seem to be a limiting
factor in network design. ResNets double the number
of feature planes per layer each time the linear size of the hidden layers
is down-scaled by a factor of two, in order to keep the computational cost per layer
constant.
However, each down-sampling halves the number of hidden units,
so most of the network's depth has to come after the input images have
been scaled down by at least a factor of 16.
Similarly, the placement
of batch-normalization blocks in Inception-IV was limited by memory
consideration \citep{journals/corr/SzegedyIV16}. Bandwidth is another
reason to take memory into account. Finite data transfer speeds and cache sizes make it harder to make full use of all available processing power.

To improve efficiency---reducing memory and power requirements---networks have been trained with low-precision floating-point arithmetic, fixed-point arithmetic, and even binary arithmetic \citep{
  nips-6:Simard+Graf:1994,
  oai:CiteSeerX.psu:10.1.1.308.2766,
  oai:arXiv.org:1412.7024,
  conf/icml/GuptaAGN15,
  DBLP:journals/corr/LinCMB15,
  journals/corr/CourbariauxBD15,
  xnor-net,
  miyashita2016convolutional,
  quant}.
These experiments have been carried out for relatively
shallow networks---generally between three and eight layers.
We will focus on much deeper networks. Well-designed deep networks are more efficient than wide, shallow networks, so we believe that reducing the level of precision is more challenging, but more useful when successful.

With this in mind, we have tried to train modern network architectures
using limited-precision arithmetic, focusing on the activations halfway through each batch-normalization module, after normalization, before the affine transform.

In Section \ref{sec:memory} we briefly explore the differences in memory consumption for neural networks at test and training time.
In Section \ref{sec:ForwardPropWithoutMul} we discuss converting complicated floating-point multiplications into simpler integer addition operations.
In Section \ref{sec:batchnorm} we recap batch-normalization, splitting it into normalization and affine transform steps.

We focus our experiments on classes of models that are popular in
computer vision. However, ideas from computer vision have informed
network design in other areas, such as speech recognition \citep{journals/corr/XiongDHSSSYZ16}
and neural machine translation \citep{oai:arXiv.org:1606.04199,journals/corr/GehringAGD16,DBLP:journals/corr/WuSCLNMKCGMKSJL16},
and have been extended to three dimensions to process video \citep{journals/corr/TranBFTP14}, so low precision batch-normalized activations should
be applicable more broadly. The networks we consider use batch-normalization after each learnt layer, before applying an activation function.
We explain this in Section \ref{sec:cromulent}.

Attempts to use lower precision arithmetic have often combined a range of levels of precision, depending on the location.
In general, lower precision is needed when storing the network weights and activations during the forward pass, while higher precision
is needed during accumulation steps, e.g. for matrix multiplication, backpropagation and gradient descent.
Storing the activations accurately seems to be crucial. From a full precision AlexNet/ImageNet top-5 accuracy baseline of 80.2\%, binarizing only the weights reduces the accuracy only marginally to 79.4\%, but binarizing the activations as well reduces the accuracy to 69.2\% \citep{xnor-net}. That is still extremely impressive performance, given the replacement of floating-point arithmetic with much simpler 1-bit XNOR operations, but it clearly cannot be used as a drop-in replacement for existing network designs. Also, it seems unlikely that binary activations are compatible with residual connections. We consider a range of quantization schemes using between two and eight bits, see Section \ref{sec:lowprec}.

\subsection{Test and training memory use}\label{sec:memory}

Neural network memory usage is very different during training and testing. At test
time, hidden units only need to be stored long enough to calculate
the values of any output nodes. For feed-forward networks such
as VGG \citep{journals/corr/SimonyanZ14a}, this means storing one hidden layer to calculate the next.
For modular networks such as ResNets \citep{ResNet}, once the output of a module
is calculated, all the internal state can be forgotten. In contrast,
DenseNets \citep{DenseNet} accumulate a relatively large state within the densely connected
blocks. The peak memory used when calculating the forward pass
for a single image may not seem huge, but it is significant compared
to, say, the size of a typical mobile processor's cache.

At training time, much more memory is needed.
Many of the activations generated during the forward pass must be stored to efficiently implement backpropagation.
Additionally, training is generally carried out with batches of size 64--256 for reasons of computational efficiency, and to benefit from batch-normalization.

Quantizing the batch-normalized activations has potential advantages both at training and
at test time, reducing memory requirements and computational burden.

\subsection{Forward Propagation with fewer multiplications}\label{sec:ForwardPropWithoutMul}
In addition to saving memory, some low precision representations also
have the useful property of allowing general floating-point multiplication
to be replaced with integer addition. In \citet{nips-6:Simard+Graf:1994}, the authors take advantage of the
finite range of the tanh activation function; they approximate the output
using only 3 bits, with the form $\pm2^{k}$. Using one of the
three bits to store the sign, the other two bits allow $k$ to have
range four, i.e. $k\in\{-3,-2,-1,0\}$. They also use a similar trick
during backpropagation but using five bits. In addition to saving memory, the special form in which the activations
are stored means that they can be multiplied by the network weights
using just integer addition in the exponent, and a 1-bit XOR for the
sign bit. This is computationally much simpler than regular floating-point
multiplication.

Unlike tanh, rectified linear units have unbounded range, so quantization may be less reliable.
Batch-normalization is a double-edged sword: it regularizes the typical range of the activation during the forward pass, but it increases the complexity of the backward pass.
For these reasons, we consider a range of quantization schemes, including log-scale ones that allow multiplications to be replaced by addition in the exponent.

\subsection{Batch-Normalization}\label{sec:batchnorm}
We will briefly review the definition of batch-normalization and its effect on backpropagation.
We will separate the normal batch-normalization layer into its two constituent parts, a feature-wise normalization N, and a feature-wise, learnt, affine transform A$_{a,b}$.
Let $x$ denote a vector of features.
During training, the vector is normalized to have mean zero and variance approximately one:
\begin{equation}
\mathrm{N}(x)=\frac{x-\mathrm{mean}(x)}{\sqrt{\mathrm{Var}(x)+\varepsilon}}\label{eq:bn}
\end{equation}
The affine transform is then applied
\[
\mathrm{BN}(x)=\mathrm{A}_{a,b}(\mathrm{N}(x))=a\cdot\mathrm{N}(x)+b.
\]
During backpropagation, the gradient is adjusted using the formula ($\nabla x \equiv \partial \mathrm{cost} /\partial x$)
\begin{equation}
\nabla x=\frac{\nabla N-\mathrm{mean}(\mathrm{\nabla N})-\mathrm{N}\ccirc\mathrm{mean}(\mathrm{N}\ccirc\nabla\mathrm{N})}{\sqrt{\mathrm{Var}(x)+\varepsilon}}\label{eq:backprop-bn}
\end{equation}
The normalized values N$(x)$ are good candidates for storing using low precision for two reasons. Having been normalized, they are easier to store efficiently as typically the bulk of the values lie within a limited range, say $[-6,6]$. Secondly, the N$(x)$ occur twice in \eqref{eq:backprop-bn}, so we need to store them in some form to apply the chain rule.

\subsection{Cromulent networks}\label{sec:cromulent}
We will focus on networks that have a specific form. Many popular network architectures either have this form, or can be modified to have it with minor changes.
Consider a network composed of the following types of modules:

\noindent-- Learnt layers such as fully-connected (Linear) layers, or convolutional
layers. For backpropagation, the input needs to be available, but the output does not.
We can also consider the identity function to be a `trivial' learnt layer.

\noindent-- Batch-Normalization: BN$_{a,b}(x)$=A$_{a,b}$(N$(x)$). N$(x)$ is needed for backpropagation through both the affine layer and the normalization layer. N$(x)$ is commonly recalculated from $x$ during the backward pass. We are proposing instead just to store a quantized version of N$(x)$.

\noindent-- The rectified-linear (ReLU) activation function. Either the input
or the output is needed for backpropagation.

\noindent-- `Branch' nodes that have one input, and produces as output multiple
identical copies of the input. Nothing needs to be stored for backpropagation; the incoming gradients are simply summed together elementwise.

\noindent-- `Add' nodes, that take multiple inputs, and add them together elementwise.
The activations do not need to be stored for backpropagation.

\noindent-- Average pooling. Nothing needs to be stored for backpropagation. Max-pooling is also allowed if the indices of the maximal inputs within each pooling region are
stored.

We will say that the network is {\em cromulent} if any forward path through the network has the form
\[
\dotsb\textrm{-}\star\textrm{-N-A}_{a,b}\textrm{-ReLU-Learnt-}\star\textrm{-N-A}_{a,b}\textrm{-ReLU-Learnt-}\star\textrm{-}\dotsb
\]
with $\star$ denoting either a direct connection, a branch node, an add node or a
pooling node. Cromulent networks include VGG-like networks; pre-activated ResNets and WideResNets; and DenseNets, with or without bottlenecks and compression. Some networks, such as Inception V4, SqueezeNets, and FractalNets are not cromulent, but can be turned into cromulent networks by converting them into pre-activated form, with BN-ReLU-Convolution blocks replacing Convolution-BN-ReLU blocks.

The motivation for our definition of cromulent is to ensure the networks permit a number of optimizations:

\noindent-- At training time, to do backpropagation it is sufficient to store the N$(x)$, as you can easily recalculate the adjacent Affine-ReLU transformations.
The recalculation is very lightweight, compared to the learnt layers and the N part of batch-normalization.

\noindent-- Combining the Affine-ReLU-Learnt layers into one so multiplications are replaced with additions; see Section \ref{sec:lowprec}.

\noindent-- For VGG and DenseNets at test time, the calculation of N$(x)$ can be rolled into the preceding learnt layer, meaning the output
of the learnt layer can be stored directly in compressed form, saving bandwidth
and memory. This is especially useful for DenseNets which accumulate
a large state within each densely connected block.

\subsection{Low precision representations}\label{sec:lowprec}

In Table \ref{tab:Approximation-formulae} we define eight quantization formulae.
The first seven are similar to ones from \citet{miyashita2016convolutional}; however, we apply the formulae immediately after normalization, before the Affine-ReLU layers, so there are three key differences.
Firstly, the input can be negative. We therefore use one bit to store the sign of the input---this seems to be more efficient overall as it save us from having to store additional information from the forward pass to apply equation \eqref{eq:backprop-bn}.
Secondly, we can fix the output scale, rather than having to vary the output range according to the level within the network as in their paper.
Thirdly, it allows a single normalization to be paired with different affine transformations, which is needed to implement DenseNets.

The first four approximations use a log-scale, using 2, 3, 4 and 5 bits of
memory respectively. They are compatible with replacing the multiplications during the forward pass with additions.
We also consider three uniform scales, using 4, 5, and 8 bits.
The uniform scales provide a different route to more efficient computations. Quantization the network weights with a uniform scale during the forward pass would allow the floating-point multiplications to be replaced with, say, 8-bit integer multiplication.

\begin{table*}[t]
\begin{centering}
  \makebox[\textwidth][c]{
    \begin{tabular}{ccl}
\hline
Scale & Bits & Formula\tabularnewline
\hline
\multirow{4}{*}{Log} & 2 & \vspace{2pt}
$\LogScale_{2}(x)=\mathrm{sign}(x)\cdot\mathrm{pow}\left(2,\nicefrac{1}{2}+\mathrm{clamp_{[-1,0]}}\left\lfloor \log_{2}|1.034x|\right\rfloor \right)$\vspace{2pt}
\tabularnewline
 & 3 & \vspace{2pt}
$\LogScale_{3}(x)=\mathrm{sign}(x)\cdot\mathrm{pow}\left(2,\mathrm{clamp_{[-1,2]}}\left\lfloor \log_{2}|1.316x|\right\rfloor \right)$\vspace{2pt}
\tabularnewline
 & 4 & \vspace{2pt}
$\LogScale_{4}(x)=\mathrm{sign}(x)\cdot\mathrm{pow}\left(2,\mathrm{clamp_{[-3,4]}}\left\lfloor \log_{2}|1.36x|\right\rfloor \right)$\vspace{2pt}
\tabularnewline
 & 5 & \vspace{2pt}
$\LogScale_{5}(x)=\mathrm{sign}(x)\cdot\mathrm{pow}\left(\sqrt{2},\mathrm{clamp_{[-6,9]}}\left\lfloor \log_{\sqrt{2}}|1.177x|\right\rfloor \right)$\vspace{2pt}
\tabularnewline
\hline
\multirow{3}{*}{Uniform} & 4 & \vspace{2pt}
$\UniformScale_{4}(x)=(\nicefrac{1}{2}+\mathrm{clamp_{[-8,7]}}\left\lfloor 2x\right\rfloor )/2$\vspace{2pt}
\tabularnewline
 & 5 & \vspace{2pt}
$\UniformScale_{5}(x)=(\nicefrac{1}{2}+\mathrm{clamp_{[-16,15]}}\left\lfloor 3x\right\rfloor )/3$\vspace{2pt}
\tabularnewline
 & 8 & \vspace{2pt}
$\UniformScale_{8}(x)=(\nicefrac{1}{2}+\mathrm{clamp_{[-128,127]}}\left\lfloor 8x\right\rfloor )/8$\vspace{2pt}
\tabularnewline
\hline
Other & 4 & \vspace{2pt}
$O_{4}(x)=\mathrm{sign}(x)\cdot[\mathrm{pow}(\alpha,\nicefrac{1}{2}+\mathrm{clamp_{[0,7]}}\left\lfloor \log_{\alpha}1+|x|\right\rfloor )-1]$, $\alpha:=1.29$\vspace{2pt}
\tabularnewline
\hline
    \end{tabular}
    }
\par\end{centering}
\caption{Low precision approximation formulae\label{tab:Approximation-formulae}:
pow$(x,y)=x^{y}$, $\left\lfloor x\right\rfloor $ is the greatest
integer less than or equal to $x$, and $\mathrm{clamp}{}_{[a,b]}(x):=\max(a,\min(b,x))$. }
\end{table*}

\subsubsection{Four-bit log-scale quantization}

First look at $\LogScale_{4}$. Like all of the approximation formulae, $\LogScale_{4}$ has been constructed so that in a
weak sense $\LogScale_{4}(x)\approx x$ for $x$ in a neighborhood of zero.
Its range has size $16=2^{4}$, consisting of the values
\[
\pm\nicefrac{1}{8},\pm\nicefrac{1}{4},\pm\nicefrac{1}{2},\pm1,\pm2,\pm4,\pm8,\pm16.
\]
In words, to calculate $\LogScale_4(x)$, we scale $x$ by a constant factor, take absolute value to make it
positive, take logs to the base 2, round down to the nearest integer,
restrict the value to the range $[-3,4]$, invert the log operation
by raising 2 to that power, and then multiply by sign$(x)=\pm1$ to
restore symmetry. The choice $[-3,4]$ for the range of the exponent
is a compromise between accuracy near zero and being able to represent
values further away from one. If we put a standard Gaussian random
variable $X\sim N(0,1)$ into $\LogScale_{4}$, then the mean value is preserved
by symmetry, and the standard deviation is preserved thanks to the
choice of the constant 1.36.
Note that for a floating-point number $x$, calculating $\left\lfloor \log_{2}x\right\rfloor$ is just a matter of inspecting the exponent.

\begin{figure}[t]
\begin{centering}
\includegraphics[width=0.9\columnwidth]{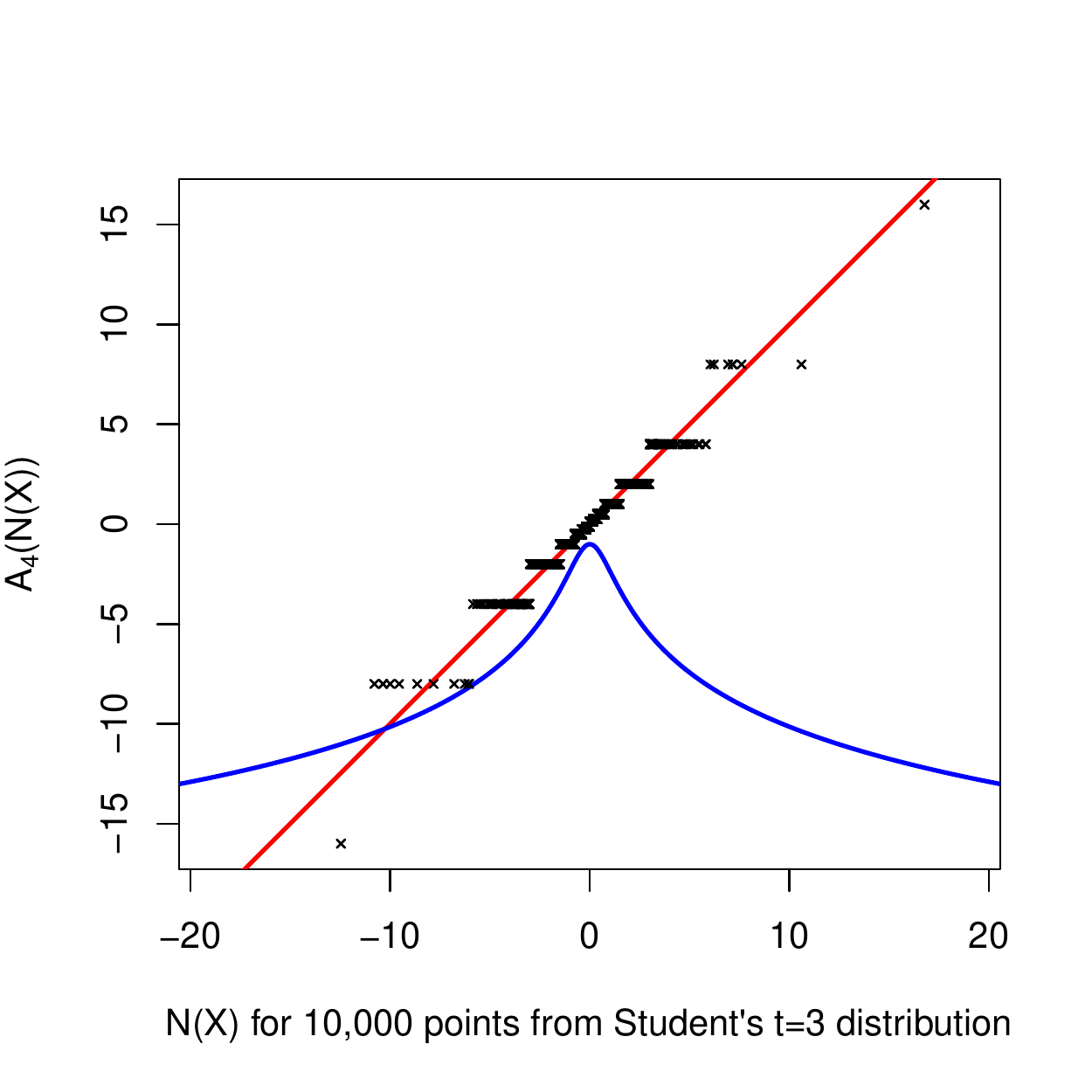}
\par\end{centering}
\caption{
  Four-bit log-scale quantization for the normalized, heavy-tailed Student's $t=3$ distribution.
  The red line is the identity function. The blue line is the log probability density function.\label{fig:Student-t}}
\end{figure}

We can look at $\LogScale_4$ as a form of dropout with
respect to all but the most significant bit of $x$.
Unlike normal dropout, it is applied in the same way during training
and testing. To measure how destructive this form of dropout is, we
can look the correlation between N$(X)$ and $\LogScale_{4}(\mathrm{N}(X))$.
See Figure \ref{fig:Student-t} and Table \ref{tab:Correlation-test-for}.
Even for a moderately heavy tailed distribution, the output of $\LogScale_{4}$ is highly correlated with the input---by Chebyshev's inequality, at least 99.6\% of the N$(x)$ must lie within the range of $\LogScale_4$.
The variance of $\LogScale_{4}($N$(x))$ will generally be close to one.

\begin{table*}[t]
\begin{centering}
\begin{tabular}{cccc}
\hline
Bits $b$ & Distribution X & Correlation$(X,\LogScale_{b}(\mathrm{N}(X)))$ & sd$(\LogScale_{b}(\mathrm{N}(X)))$\tabularnewline
\hline
2 & N(0,1) & 0.918 & 1.000\tabularnewline
3 & N(0,1) & 0.965 & 1.000\tabularnewline
4 & N(0,1) & 0.981 & 1.000\tabularnewline
2 & Student's t=3 & 0.769 & 0.888\tabularnewline
3 & Student's t=3 & 0.857 & 1.11\tabularnewline
4 & Student's t=3 & 0.970 & 0.978\tabularnewline
\hline
\end{tabular}
\par\end{centering}
\caption{Correlation and variance tests for $\LogScale_{2},$ $\LogScale_{3}$ and $\LogScale_{4}$
for a standard Gaussian and a heavy tailed probability distribution.\label{tab:Correlation-test-for}}
\end{table*}

Replacing multiplications with additions in the exponent requires
a little algebra to merge together the Affine-ReLU-Learnt triplet of layers.
Let N$(x)_{i}$ denote the $i$-th element of the the output
of the normalization operation, let $a$ and $b$ denote the parameters
of the affine transform, and let $w$ denote a network parameter.
Then
\begin{align*}
&\text{ReLU}(\text{BN}(x)_{i})\cdot w=\\
&\quad\begin{cases}
(aw)\LogScale_{4}(\text{N}(x)_{i})+(bw) & \textrm{if }\LogScale_{4}(\text{N}(x)_{i})>-b/a,\\
0 & \textrm{otherwise}.
\end{cases}
\end{align*}
We have replaced a number of floating-point multiplications for integer
addition and floating-point addition. (Alternatively, this could be
implemented with bit-shifting and fixed-point addition).
This is significant as it would allow networks to be deployed to less powerful, more energy efficient devices.

During training, we can save a substantial amount of memory by using $\LogScale_4(\text{N}(x))$ in place of N$(x)$  in equation \eqref{eq:backprop-bn} and, and by recalculating the Affine-ReLU transformation as needed, rather than saving the result.

\begin{figure*}[h]
\begin{centering}
\makebox[\textwidth][c]{\includegraphics[width=0.8\columnwidth]{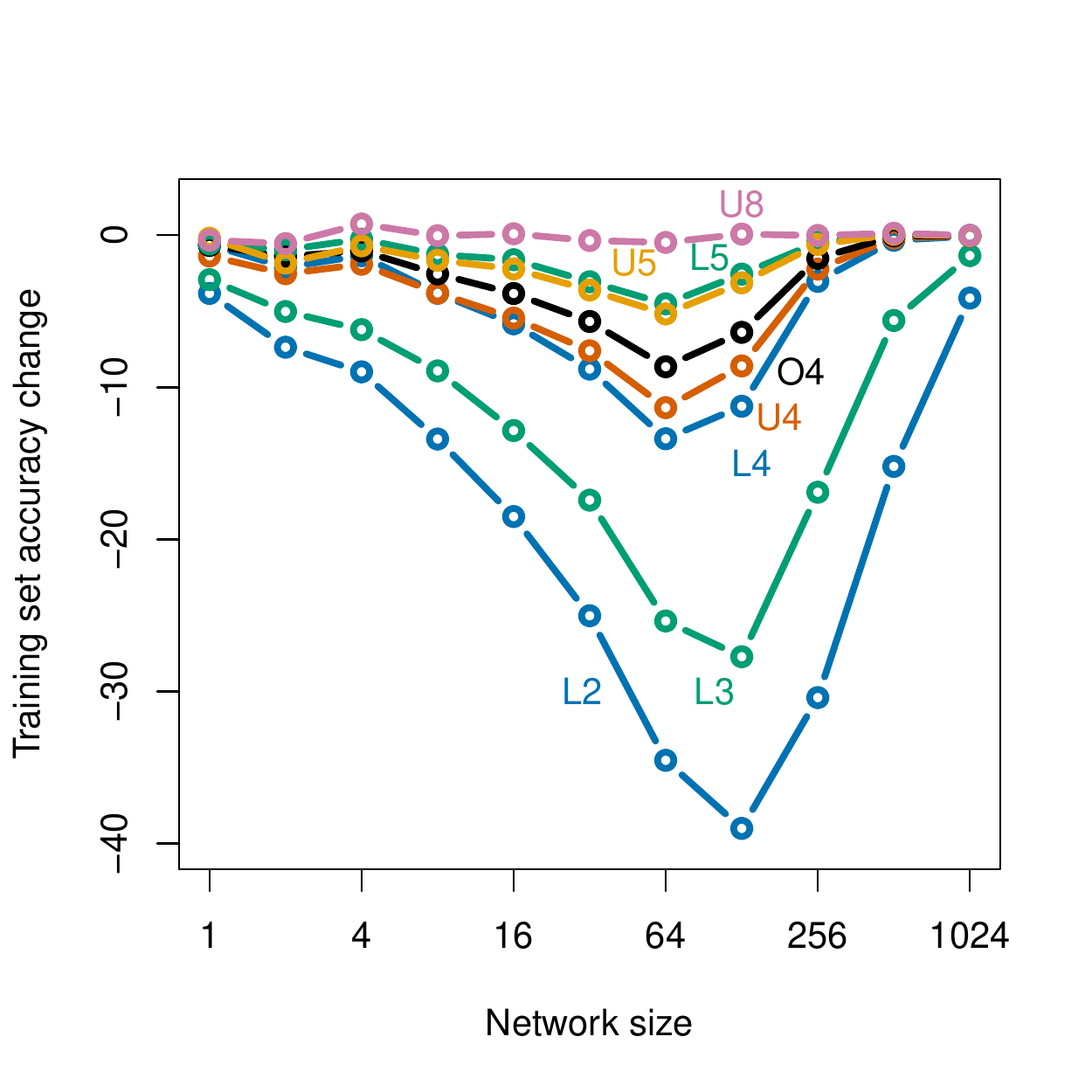}\includegraphics[width=0.8\columnwidth]{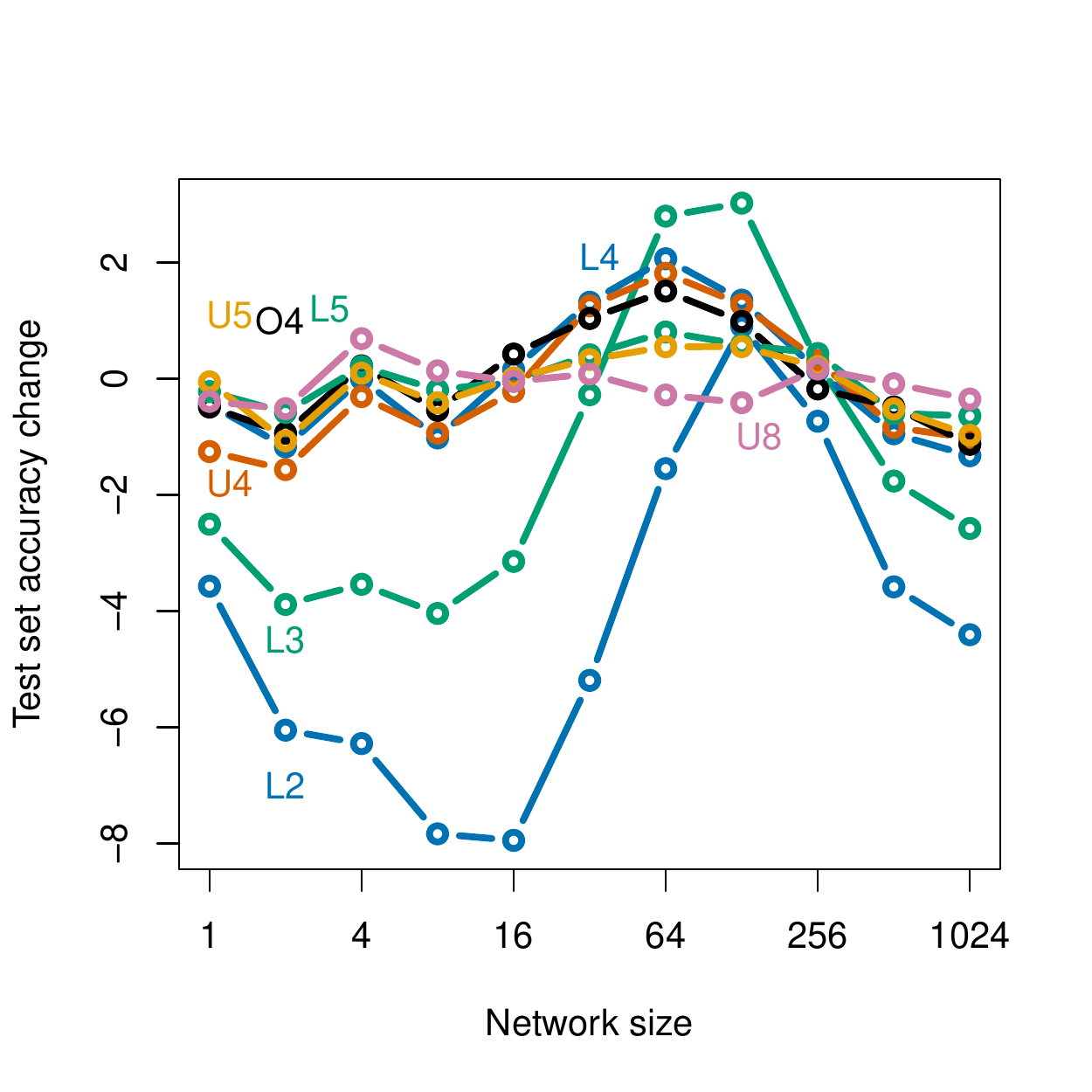}}
\par\end{centering}
\caption{Percentage point changes in the accuracy on the training and test sets relative to regular batch-normalization. Baseline test accuracy increases from 19.5\% ($N=1$) to 48.1\% ($N=32$) to 58.7\% ($N=1024$).
Note that the two $y$-axes use different scales. \label{fig:CIFAR-FC}}
\end{figure*}

\subsubsection{Other representations}

We have also defined a variety of other functions corresponding to
low precision representations. Using three bits we can represent a
subset of the range of $\LogScale_{4}$, specifically $\pm\nicefrac{1}{2},\pm1,\pm2,\pm4$.
Using two bits, we can represent the values $\pm1/\sqrt{2},\pm\sqrt{2}$.
We can still perform the low precision multiplication trick, by storing
the weights multiplied by $\sqrt{2}$.
With 5 bits, we can increase the precision by using a log-scale with
base $\sqrt{2}$; the multiplication trick can be modified by storing
$w$ and $w\sqrt{2}$.

To construct the uniform scales, we had to compromise between accuracy and width of support.
The final representation in Table \ref{tab:Approximation-formulae} is an attempt to improve accuracy at the expense of simplicity. The set of output points is less concentrated around zero, allowing it to provide better
accuracy further afield. The number 1.29 is chosen to give a range
of approximately $[-6,6]$.

\section{Experiments}

We have performed a range of experiments with the CIFAR-10 and ImageNet datasets.

\subsection{Fully connected, permutation invariant CIFAR-10}

To test low precision approximations for fully-connected networks, we treated
CIFAR-10's $3\times32\times32$ images as unordered collections of
3072 features, without data augmentation, and preprocessed only by
scaling to the interval $[-1,1]$. We constructed a range of two layer,
fully connected networks,
\[
\mathrm{FC}(2^{n})-\mathrm{BN}-\mathrm{ReLU}-\mathrm{FC}(2^{n})-\mathrm{BN}-\mathrm{ReLU}-\mathrm{FC}(10)
\]
with $n=0,1,\dots,10$, for each of the quantization schemes. We
trained the networks for 100 epochs in batches of size 100, with learning
rate $10^{-2}$ and Nesterov momentum of 0.9. See Figure \ref{fig:CIFAR-FC}. The two and three bit approximation
schemes seem to have a strong regularizing effect during training,
with substantially worse recall of the training labels. On the test
set the effect is much smaller, and even positive for networks of
size 64 to 128, suggesting the some overfitting is being prevented.
This suggests that very low precision storage could be useful as a
computationally more efficient alternative to regular batch-normalization combined with dropout.

\begin{table*}[t]
\setlength{\tabcolsep}{2pt}
\begin{centering}
  \makebox[\textwidth][c]{
\begin{tabular}{cccccccc}
\hline
Network & VGG(32) & VGG(64) & ResNet-20 & ResNet-44 & WideResNet-(4,40) & DenseNet-(40,12) & DenseNet-(100,12)\tabularnewline
\hline
fp32             & 7.81  & 6.45  & 8.14  & 6.71  & 4.54  & 5.29  & 4.26 \tabularnewline
\hline
$\LogScale_{2}$     &26.23  &22.24  & 58.48 & -     & 55.99 & 48.65  & -\tabularnewline
$\LogScale_{3}$     & 9.98  & 7.42  & 11.41 & 10.13 & 6.00  & 7.43  & 5.41 \tabularnewline
$\LogScale_{4}$     & 8.84  & 7.00  & 9.52  & 7.80  & 4.81  & 6.11  & 4.64 \tabularnewline
$\LogScale_{5}$     & 8.01  & 6.44  & 8.20  & 6.95  & 4.51  & 5.45  & 4.29 \tabularnewline
\hline
$\UniformScale_{4}$ & 8.19  & 6.52  & 8.41  & 7.47  & 4.65  & 5.89  & 4.52 \tabularnewline
$\UniformScale_{5}$ & 7.99  & 6.63  & 8.12  & 6.86  & 4.48  & 5.27  & 4.24 \tabularnewline
$\UniformScale_{8}$ & 7.95  & 6.42  & 7.84  & 6.80  & 4.52  & 5.41  & 4.11 \tabularnewline
\hline
$O_{4}$             & 8.17  & 6.72  & 8.38  & 7.22  & 4.58  & 5.51  & 4.35 \tabularnewline
\hline
\end{tabular}
}
\par\end{centering}
\setlength{\tabcolsep}{6pt}
\caption{Results for convolutional networks for CIFAR-10. Repeated five times,
we report the mean percentage error rates.}
\end{table*}

\subsection{CIFAR-10}\label{cifar10conv}

Using the fb.resnet.torch\footnote{\url{https://github.com/facebook/fb.resnet.torch}}
package we trained a variety of convolutional networks on CIFAR-10.
We trained two VGG-like \citep{journals/corr/SimonyanZ14a} network
with pairs of $3\times3$ convolutions, followed by $3\times3$ max-pooling
with stride 2, and the number of features doubling after each max-pooling halves the spatial size, i.e.
\begin{align*}
 & M(k)-M(2k)-M(4k)-(8k)\mathrm{C}4\textrm{ with}\\
 & M(k):=[k\textrm{C3-BN-ReLU-}k\textrm{C3-MP3/2-BN-ReLU]}
\end{align*}
with $k=32$ and $k=64$. These are relatively shallow networks, but are included for the sake of comparison.

To experiment with deeper networks, we also trained ResNet-20 and
ResNet-44 networks \citep{ResNet}, WideResNets \citep{Zagoruyko2016WRN}
with depth 40 and widening factor 4, and DenseNets \citep{DenseNet}
with growth rate $k=12$ and depths 40 and 100. We used the default
learning rate schedule for the VGG and ResNet networks: 164 epochs
with learning rate annealing. For WideResNets and DenseNets, we used their
customary 200 and 300 epoch learning rate schedules, respectively.
All networks were trained with weight decay of $10^{-4}$, and data
augmentation in the form of random crops after 4-pixel zero padding,
and horizontal flips.

$\LogScale_{2}$ clearly inhibits proper training, at least with the defaults
learning meta-parameters. $\LogScale_{3}$ and $\LogScale_{4}$ perform much better,
especially when you consider the reduction in computational complexity
of the operations involved. In this set of experiments, the uniform scale quantizations are broadly similar to the log-scale ones.
Although not computationally simpler, the $O_{4}$ results show that substantial reductions can be
made in memory usage with minimal loss of accuracy.

\begin{table*}[t]
\begin{centering}
\begin{tabular}{cccccccccc}
\hline
Network &\quad& \multicolumn{2}{c}{ResNet-18} &\quad & \multicolumn{2}{c}{ResNet-50 } &\quad& \multicolumn{2}{c}{WideResNet-(2,18) }\tabularnewline
\hline
fp32                && 30.43 & 10.76 && 24.01 &  7.02 && 25.94 &  8.19  \tabularnewline
\hline
$\LogScale_{3}$     && 36.45 & 14.83 && 43.88 & 20.39 && 29.93 & 10.71  \tabularnewline
$\LogScale_{4}$     && 33.28 & 12.54 && 26.19 &  8.16 && 27.24 &  9.01  \tabularnewline
$\LogScale_{5}$     && 31.05 & 11.45 && 24.39 &  7.39 && 25.96 &  8.31  \tabularnewline
\hline
$\UniformScale_{4}$ && 36.55 & 15.02 && 49.75 & 25.82 && 30.00 & 10.92  \tabularnewline
$\UniformScale_{5}$ && 30.72 & 11.24 && 34.72 & 13.67 && 26.03 &  8.29  \tabularnewline
$\UniformScale_{8}$ && 30.15 & 10.73 && 25.68 &  7.93 && 25.95 &  8.21  \tabularnewline
\hline
$O_{4}$             && 31.67 & 11.41 && 27.27 &  8.81 && 26.03 &  8.37  \tabularnewline
\hline
\end{tabular}
\par\end{centering}
\caption{ImageNet top-1 and top-5 validation error rates. The first batch-normalization operation of each network is not quantized.\label{tbl:imagenet} }
\end{table*}

\subsection{ImageNet}

Again using fb.resnet.torch, we trained a number of ResNet models on ImageNet.
Models are trained for 90 epochs with weight decay $10^{-4}$, except where noted.

The ResNet-18 consists of 8 basic residual blocks; ResNet-50 consists of 14 bottleneck residual blocks. We also trained a wide ResNet with widening factor 2, depth 18; all layers after the initial convolution have double the width of a regular ResNet-18. Although the bodies of these networks are cromulent, they begin with modules
\begin{equation}\label{notCromulent}
64\text{C}7/2-\text{BN}-\text{ReLU}-\text{MP}3/2,
\end{equation}
mapping the 3x224x224 input to size 64x56x56. To avoid changing the model definition, we decided to keep the initial BN operation unchanged, using low precision BN for all subsequent BN layers. See Table \ref{tbl:imagenet}.

Here we find that the uniform scale quantization can inhibit learning, while the other representations do better.
Perhaps unsurprisingly, quantization affects the deep ResNet-50 more than the WideResNet-18.
The failure of the uniform quantization methods are in stark contrast to the results for CIFAR-10, demonstrating that quantization methods cannot be assumed to be `mostly-harmless' without extensive testing on challenging datasets.

The $O_4$ version of ResNet-18 has a top-5 error 0.65 percentage points higher than the 32-bit baseline. To see if performance was suffering due to the use of weight decay, we reduced the  decay rate to $10^{-5}$ and trained for ten additional epochs: the top-1/top5 errors improved to 31.15\% / 11.13\%, narrowing the gap to 0.37\%. This suggests that the low precision models need less regularization, but are capable of very similar levels of performance.

\subsection{Mixed networks with fine-tuning}

An alternative to training networks with reduced precision arithmetic
is to take a network trained using 32-bit precision, and then try to compress the network to reduce test-time complexity, for example by pruning small weights, and quantizing the remaining weights \citep{journals/corr/HanMD15}. With this in mind, we try substituting
regular batch-normalization for $\LogScale_4$-BN and $\LogScale_5$-BN in pretrained DenseNet-121 and DenseNet-201 ImageNet networks.

We tried quantizing (i) all normalized activations, and (ii) all but the first normalization, and (iii) all the normalizations in the second, third and fourth densely connected blocks (all but the first 14 BN-operations). See Tables \ref{tbl:retrofit} and \ref{tbl:retrofit2}. The early layers of the network, where the number of features planes is small, seem to be most sensitive to reductions in accuracy.

We did the same with a ResNet-18. For this experiment we made the network fully cromulent: with reference to \ref{notCromulent}, we moved the max-pooling module to immediately after the first convolution. We tried quantizing all the normalization operations, and all but the first normalization. Again, except at the beginning of the network, quantizing the normalized activations preserves most of their functionality.

\begin{table*}[t]
\begin{centering}
\begin{tabular}{lcccc}
\hline
Network & \multicolumn{2}{c}{Raw} & \multicolumn{2}{c}{Fine-tuned}\tabularnewline
\hline
DenseNet-121                            & 25.25 &  7.88 & -     & -    \tabularnewline
\ ...quantizing all 121 BN              & 32.05 & 12.07 & 27.33 & 9.04 \tabularnewline
\ ...quantizing all but the first BN              & 30.87 & 11.13 & 27.11 & 8.87 \tabularnewline
\ ...quantizing 3 out of 4 dense blocks & 27.97 &  9.34 & 26.42 & 8.38 \tabularnewline
\hline
DenseNet-201                            & 22.78 & 6.46 & -      & -    \tabularnewline
\ ...quantizing all 201 BN              & 28.61 & 9.87 & 25.00  & 7.54 \tabularnewline
\ ...quantizing all but the first BN          & 27.97 & 9.21 & 24.94  & 7.33 \tabularnewline
\ ...quantizing 3 out of 4 dense blocks & 25.42 & 7.68 & 24.06  & 7.07 \tabularnewline
\hline
ResNet-18                               & 30.21 & 10.79 & -     & -     \tabularnewline
\ ...quantizing all BN                  & 36.26 & 14.72 & 33.71 & 12.44 \tabularnewline
\ ...quantizing all but first BN        & 34.58 & 13.40 & 32.16 & 12.03 \tabularnewline
\hline
\end{tabular}
\par\end{centering}
\caption{Fine-tuning a DenseNet-121/201/ResNet18, retrofitted with $\LogScale_4$ low-precision storage,
for ten epochs with weight decay $10^{-5}$.\label{tbl:retrofit}}
\begin{centering}
\begin{tabular}{lcccc}
\hline
Network & \multicolumn{2}{c}{Raw} & \multicolumn{2}{c}{Fine-tuned}\tabularnewline
\hline
DenseNet-121                            & 25.25 &  7.88 & -     & -    \tabularnewline
\ ...quantizing all 121 BN              & 26.69 &  8.65 & 25.32 & 7.94 \tabularnewline
\ ...quantizing all but the first BN              & 26.26 &  8.48 & 25.28 & 7.88 \tabularnewline
\ ...quantizing 3 out of 4 dense blocks & 25.75 &  8.20 & 25.21 & 7.72 \tabularnewline
\hline
DenseNet-201                            & 22.78 & 6.46 &-      & -    \tabularnewline
\ ...quantizing all 201 BN              & 23.84 & 7.16 & 23.47 & 6.69 \tabularnewline
\ ...quantizing all but the first  BN          & 23.71 & 6.90 & 23.33 & 6.59 \tabularnewline
\ ...quantizing 3 out of 4 dense blocks & 23.35 & 6.71 & 23.18 & 6.59 \tabularnewline
\hline
ResNet-18                               & 30.21 & 10.79 & -     & -   \tabularnewline
\ ...quantizing all BN                  & 31.53 & 11.52 & 30.63 & 11.00 \tabularnewline
\ ...quantizing all but first BN        & 31.06 & 11.32 & 30.42 & 11.00 \tabularnewline
\hline
\end{tabular}
\par\end{centering}
\caption{Fine-tuning a DenseNet-121/201/ResNet18, retrofitted with $\LogScale_5$ low-precision storage,
for one epoch.\label{tbl:retrofit2}}
\end{table*}

\subsection{Training with quantized gradients}
We have quantized the activations; we have not so far quantized the gradients---the partial derivatives of the cost function with respect to the activations. While \citet{quant,miyashita2016convolutional} have shown this is possible for feed-forward networks, the situation is more complicated for network architectures with multiple connections such as DenseNets. Consider the DenseNet-$(3N+4,12)$ networks from Section~\ref{cifar10conv} ($N=12$ and $32$). Each of the three densely connected blocks consists of an input followed by $N$ convolutions. The $n$-th convolution takes the original input concatenated with the outputs of the $n-1$ preceding convolutions. During the backward pass, it has to somehow send the backpropagation error signal to each of its $n$ inputs.

The messages cannot realistically be sent directly; the number of messages would grow with order $N^2$. Instead, the messages are accumulated additively working backwards through the dense block\footnote{\url{https://github.com/liuzhuang13/DenseNet/}}.

Quantizing the gradients corresponds to iteratively quantizing partial sums of the messages. Experimentally, this amplifies numerical imprecision, destroying the capacity of the DenseNets to learn CIFAR-10.

Alternatively, the gradients can be backpropagated with floating-point precision, but quantized immediately before being applied to each convolution---converting multiplications to additions, but not reducing the overall memory footprint. We trained $\LogScale_5$-quantized DenseNets, quantizing the gradients to 8-bits using the set \mbox{$\{\pm 2^{k/2}: k=-127,\dots,-1, 0\}$}.
Compared to totally unquantized gradients, mean test errors increased very slightly: from 5.45\% to 5.64\% for $N=12$, and from 4.29\% to 4.44\% for $N=32$.

\section{Implementation}
We have looked at the experimental properties of some low precision
approximation schemes. The natural way to take advantage of this method is:

\noindent-- At training time, combine the Affine-ReLU-Convolution operations into one function/kernel, so that the memory-reads can be replaced with low precision versions.
This could be particularly useful for DenseNets, as the convolutions (or bottleneck projections) tend to have much larger input than output.

\noindent-- At test time, for VGG and DenseNets, also roll the N part of batch-normalization into the Affine-ReLU-Convolution operation, and write the output directly in quantized form.

Implementing this efficiently is a moderate engineering challenge---the same is true for all BLAS and convolutional operations. Particularly in the case of DenseNets, being able to read the input from memory in a compressed form is an advantage, as each convolutional layer takes many more activations as input than it produces as output. Storing the 4D hidden layers tensors in the form height$\times$width$\times$batch$\times$features, rather than the usual batch$\times$features$\times$height$\times$width could be helpful in terms of efficiently reading contiguous blocks from memory.

\section{Conclusion}

Modern, efficient, deep neural-networks can be trained with substantially less memory, and nearly all of the multiplication operations can be replaced with additions, all with only very minor loss in accuracy. Moreover, this technique will allow larger networks to be trained in situations where memory is currently a limiting factor, such as 3D convolutional networks for video, and large language models.

The first few layers of a network, where the number of feature planes is smaller, seem to be more sensitive to quantization. Mixing the number of bits used, with 5 bits in the lower layers, and 3 or 4 bits in the higher layers might be a good way to balance accuracy and memory footprint.

Our experiments have been limited to ConvNets designed for use with floating-point arithmetic, using the existing training procedures without modification.
Better results may be possible by designing new network architectures, or by fine-tuning the training procedures.

Our definition of {\em cromulent} includes a wide range of networks. However, it is not meant to exhaustively identify every opportunity for low-precision batch-normalized activations. For example, in \citet{cooijmans2016recurrent}, batch-normalization is added to an LSTM recurrent network. LSTM cells are built using sigmoid and tanh activations instead of ReLUs, but much of the internal state could still potentially be stored at reduced precision.

To keep our analysis focused, we have not tried to optimize everything at the same time, focusing instead on just the activations.
There are almost certainly additional opportunities to improve the computational performance of networks with low-precision batch-normalized activations:

\noindent-- ConvNets are often robust with respect to quantizing the network weights \citep{DBLP:journals/corr/MerollaAAEM16}. Allowing the network weights to take just three values, a ResNet-18 achieved 15.8\% validation error on ImageNet \citep{journals/corr/LiL16}. An interesting problem for future work is merging mild weight quantization with low-precision batch-normalized-activations whilst minimizing loss of accuracy.

\noindent-- If the activations and network weights are quantized, lower-precision arithmetic might be sufficient for summing the activation$\times$weight terms for each convolutional layer.

Some of these optimizations will be immediately useful, particularly the reduced memory overhead. To take full advantage of these techiques will require the development of new, low-power hardware.
\bibliography{arxiv}

\begin{thebibliography}{27}
\providecommand{\natexlab}[1]{#1}
\providecommand{\url}[1]{\texttt{#1}}
\expandafter\ifx\csname urlstyle\endcsname\relax
  \providecommand{\doi}[1]{doi: #1}\else
  \providecommand{\doi}{doi: \begingroup \urlstyle{rm}\Url}\fi

\bibitem[Cooijmans et~al.(2016)Cooijmans, Ballas, Laurent, G{\"u}l{\c{c}}ehre,
  and Courville]{cooijmans2016recurrent}
T.~Cooijmans, N.~Ballas, C.~Laurent, {\c{C}}.~G{\"u}l{\c{c}}ehre, and
  A.~Courville.
\newblock Recurrent batch normalization.
\newblock \emph{arXiv preprint arXiv:1603.09025}, 2016.
\newblock URL \url{http://arxiv.org/abs/1603.09025}.

\bibitem[Courbariaux et~al.(2014)Courbariaux, Bengio, and
  David]{oai:arXiv.org:1412.7024}
M.~Courbariaux, Y.~Bengio, and J.-P. David.
\newblock Training deep neural networks with low precision multiplications,
  2014.
\newblock URL \url{http://arxiv.org/abs/1412.7024}.
\newblock Accepted as a workshop contribution at ICLR 2015.

\bibitem[Courbariaux et~al.(2015)Courbariaux, Bengio, and
  David]{journals/corr/CourbariauxBD15}
M.~Courbariaux, Y.~Bengio, and J.-P. David.
\newblock Binaryconnect: Training deep neural networks with binary weights
  during propagations.
\newblock \emph{Advances in Neural Information Processing Systems 28}, 2015.
\newblock URL \url{https://arxiv.org/abs/1511.00363}.

\bibitem[Gehring et~al.(2016)Gehring, Auli, Grangier, and
  Dauphin]{journals/corr/GehringAGD16}
J.~Gehring, M.~Auli, D.~Grangier, and Y.~N. Dauphin.
\newblock A convolutional encoder model for neural machine translation.
\newblock 2016.
\newblock URL \url{http://arxiv.org/abs/1611.02344}.

\bibitem[Gupta et~al.(2015)Gupta, Agrawal, Gopalakrishnan, and
  Narayanan]{conf/icml/GuptaAGN15}
S.~Gupta, A.~Agrawal, K.~Gopalakrishnan, and P.~Narayanan.
\newblock Deep learning with limited numerical precision.
\newblock \emph{Proceedings of the 32nd International Conference on Machine
  Learning, {ICML}}, 2015.
\newblock URL \url{http://jmlr.org/proceedings/papers/v37/}.

\bibitem[Han et~al.(2016)Han, Mao, and Dally]{journals/corr/HanMD15}
S.~Han, H.~Mao, and W.~J. Dally.
\newblock Deep compression: Compressing deep neural network with pruning,
  trained quantization and huffman coding.
\newblock \emph{ICLR}, 2016.
\newblock URL \url{http://arxiv.org/abs/1510.00149}.

\bibitem[He et~al.(2016)He, Zhang, Ren, and Sun]{ResNet}
K.~He, X.~Zhang, S.~Ren, and J.~Sun.
\newblock Identity mappings in deep residual networks.
\newblock \emph{ECCV}, 2016.
\newblock URL \url{https://arxiv.org/abs/1603.05027}.

\bibitem[Huang et~al.(2016)Huang, Liu, Weinberger, and van~der
  Maaten]{DenseNet}
G.~Huang, Z.~Liu, K.~Q. Weinberger, and L.~van~der Maaten.
\newblock Densely connected convolutional networks.
\newblock 2016.
\newblock URL \url{https://arxiv.org/abs/1608.06993}.

\bibitem[Hubara et~al.(2016)Hubara, Courbariaux, Soudry, El-Yaniv, and
  Bengio]{quant}
I.~Hubara, M.~Courbariaux, D.~Soudry, R.~El-Yaniv, and Y.~Bengio.
\newblock Quantized neural networks: Training neural networks with low
  precision weights and activations.
\newblock 2016.
\newblock URL \url{http://arxiv.org/pdf/1609.07061}.

\bibitem[Iandola et~al.(2016)Iandola, Han, Moskewicz, Ashraf, Dally, and
  Keutzer]{oai:arXiv.org:1602.07360}
F.~N. Iandola, S.~Han, M.~W. Moskewicz, K.~Ashraf, W.~J. Dally, and K.~Keutzer.
\newblock Squeezenet: Alexnet-level accuracy with 50x fewer parameters and
  $<$0.5{MB} model size, 2016.
\newblock URL \url{http://arxiv.org/abs/1602.07360}.

\bibitem[Ioffe and Szegedy(2015)]{BatchNorm}
S.~Ioffe and C.~Szegedy.
\newblock Batch normalization: Accelerating deep network training by reducing
  internal covariate shift.
\newblock \emph{Proceedings of the 32nd International Conference on Machine
  Learning}, 2015.
\newblock URL \url{http://jmlr.org/proceedings/papers/v37/ioffe15.pdf}.

\bibitem[Larsson et~al.(2016)Larsson, Maire, and
  Shakhnarovich]{larsson2016fractalnet}
G.~Larsson, M.~Maire, and G.~Shakhnarovich.
\newblock Fractal{N}et: Ultra-deep neural networks without residuals.
\newblock 2016.
\newblock URL \url{https://arxiv.org/abs/1605.07648}.

\bibitem[Li and Liu(2016)]{journals/corr/LiL16}
F.~Li and B.~Liu.
\newblock Ternary weight networks.
\newblock 2016.
\newblock URL \url{http://arxiv.org/abs/1605.04711}.

\bibitem[Lin et~al.(2016)Lin, Courbariaux, Memisevic, and
  Bengio]{DBLP:journals/corr/LinCMB15}
Z.~Lin, M.~Courbariaux, R.~Memisevic, and Y.~Bengio.
\newblock Neural networks with few multiplications.
\newblock \emph{ICLR}, 2016.
\newblock URL \url{http://arxiv.org/abs/1510.03009}.

\bibitem[Merolla et~al.(2016)Merolla, Appuswamy, Arthur, Esser, and
  Modha]{DBLP:journals/corr/MerollaAAEM16}
P.~Merolla, R.~Appuswamy, J.~V. Arthur, S.~K. Esser, and D.~S. Modha.
\newblock Deep neural networks are robust to weight binarization and other
  non-linear distortions.
\newblock 2016.
\newblock URL \url{http://arxiv.org/abs/1606.01981}.

\bibitem[Miyashita et~al.(2016)Miyashita, Lee, and
  Murmann]{miyashita2016convolutional}
D.~Miyashita, E.~H. Lee, and B.~Murmann.
\newblock Convolutional neural networks using logarithmic data representation.
\newblock \emph{arXiv preprint arXiv:1603.01025}, 2016.
\newblock URL \url{http://arxiv.org/abs/1603.01025}.

\bibitem[Nair and Hinton(2010)]{conf/icml/NairH10}
V.~Nair and G.~E. Hinton.
\newblock Rectified linear units improve restricted boltzmann machines.
\newblock \emph{Proceedings of the 27th International Conference on Machine
  Learning}, 2010.
\newblock URL \url{http://www.icml2010.org/papers/432.pdf}.

\bibitem[Rastegari et~al.(2016)Rastegari, Ordonez, Redmon, and
  Farhadi]{xnor-net}
M.~Rastegari, V.~Ordonez, J.~Redmon, and A.~Farhadi.
\newblock X{N}{O}{R}-{N}et: Imagenet classification using binary convolutional
  neural networks.
\newblock \emph{ECCV}, 2016.
\newblock URL \url{http://arxiv.org/abs/1603.05279}.

\bibitem[Simard and Graf(1994)]{nips-6:Simard+Graf:1994}
P.~Y. Simard and H.~P. Graf.
\newblock Backpropagation without multiplication.
\newblock In J.~D. Cowan, G.~Tesauro, and J.~Alspector, editors, \emph{Advances
  in Neural Information Processing Systems}, volume~6, pages 232--239. Morgan
  Kaufmann Publishers, Inc., 1994.
\newblock URL
  \url{https://papers.nips.cc/paper/833-backpropagation-without-multiplication}.

\bibitem[Simonyan and Zisserman(2014)]{journals/corr/SimonyanZ14a}
K.~Simonyan and A.~Zisserman.
\newblock Very deep convolutional networks for large-scale image recognition.
\newblock 2014.
\newblock URL \url{http://arxiv.org/abs/1409.1556}.

\bibitem[Szegedy et~al.(2016)Szegedy, Ioffe, and
  Vanhoucke]{journals/corr/SzegedyIV16}
C.~Szegedy, S.~Ioffe, and V.~Vanhoucke.
\newblock Inception-v4, inception-resnet and the impact of residual connections
  on learning.
\newblock 2016.
\newblock URL \url{http://arxiv.org/abs/1602.07261}.

\bibitem[Tran et~al.(2015)Tran, Bourdev, Fergus, Torresani, and
  Paluri]{journals/corr/TranBFTP14}
D.~Tran, L.~Bourdev, R.~Fergus, L.~Torresani, and M.~Paluri.
\newblock Learning spatiotemporal features with 3d convolutional networks.
\newblock In \emph{2015 IEEE International Conference on Computer Vision
  (ICCV)}, pages 4489--4497. IEEE, 2015.
\newblock URL \url{https://arxiv.org/abs/1412.0767}.

\bibitem[Vanhoucke et~al.(2011)Vanhoucke, Senior, and
  Mao]{oai:CiteSeerX.psu:10.1.1.308.2766}
V.~Vanhoucke, A.~Senior, and M.~Z. Mao.
\newblock Improving the speed of neural networks on cpus.
\newblock In \emph{Deep Learning and Unsupervised Feature Learning Workshop,
  NIPS 2011}, 2011.

\bibitem[Wu et~al.(2016)Wu, Schuster, Chen, Le, Norouzi, Macherey, Krikun, Cao,
  Gao, Macherey, Klingner, Shah, Johnson, Liu, Kaiser, Gouws, Kato, Kudo,
  Kazawa, Stevens, Kurian, Patil, Wang, Young, Smith, Riesa, Rudnick, Vinyals,
  Corrado, Hughes, and Dean]{DBLP:journals/corr/WuSCLNMKCGMKSJL16}
Y.~Wu, M.~Schuster, Z.~Chen, Q.~V. Le, M.~Norouzi, W.~Macherey, M.~Krikun,
  Y.~Cao, Q.~Gao, K.~Macherey, J.~Klingner, A.~Shah, M.~Johnson, X.~Liu,
  L.~Kaiser, S.~Gouws, Y.~Kato, T.~Kudo, H.~Kazawa, K.~Stevens, G.~Kurian,
  N.~Patil, W.~Wang, C.~Young, J.~Smith, J.~Riesa, A.~Rudnick, O.~Vinyals,
  G.~Corrado, M.~Hughes, and J.~Dean.
\newblock Google's neural machine translation system: Bridging the gap between
  human and machine translation.
\newblock 2016.
\newblock URL \url{http://arxiv.org/abs/1609.08144}.

\bibitem[Xiong et~al.(2016)Xiong, Droppo, Huang, Seide, Seltzer, Stolcke, Yu,
  and Zweig]{journals/corr/XiongDHSSSYZ16}
W.~Xiong, J.~Droppo, X.~Huang, F.~Seide, M.~Seltzer, A.~Stolcke, D.~Yu, and
  G.~Zweig.
\newblock The {M}icrosoft 2016 conversational speech recognition system.
\newblock 2016.
\newblock URL \url{http://arxiv.org/abs/1609.03528}.

\bibitem[Zagoruyko and Komodakis(2016)]{Zagoruyko2016WRN}
S.~Zagoruyko and N.~Komodakis.
\newblock Wide residual networks.
\newblock In \emph{BMVC}, 2016.
\newblock URL \url{https://arxiv.org/abs/1605.07146}.

\bibitem[Zhou et~al.(2016)Zhou, Cao, Wang, Li, and
  Xu]{oai:arXiv.org:1606.04199}
J.~Zhou, Y.~Cao, X.~Wang, P.~Li, and W.~Xu.
\newblock Deep recurrent models with fast-forward connections for neural
  machine translation.
\newblock \emph{TACL}, 2016.
\newblock URL \url{http://arxiv.org/abs/1606.04199}.

\end{thebibliography}

\end{document}